# Exploring Sentiment Dynamics and Predictive Behaviors in Cryptocurrency Discussions by Few-Shot Learning with Large Language Models

Moein Shahiki Tash[1,*], Zahra Ahani[1], Mohim Tash[2], Olga Kolesnikova[1], and Grigori Sidorov[1]

[1]*Instituto Politécnico Nacional (IPN), Centro de Investigación en Computación (CIC), Mexico City, Mexico*
[2]*Entrepreneurship Department, Management and Economics Faculty University of Sistan and Baluchestan, Zahedan, Iran*

**Abstract**

This study performs analysis of Predictive statements, Hope speech, and Regret Detection behaviors within cryptocurrency-related discussions, leveraging advanced natural language processing techniques. We introduce a novel classification scheme named "Prediction statements," categorizing comments into Predictive Incremental, Predictive Decremental, Predictive Neutral, or Non-Predictive categories. Employing GPT-4o, a cutting-edge large language model, we explore sentiment dynamics across five prominent cryptocurrencies: Cardano, Binance, Matic, Fantom, and Ripple. Our analysis reveals distinct patterns in predictive sentiments, with Matic demonstrating a notably higher propensity for optimistic predictions. Additionally, we investigate hope and regret sentiments, uncovering nuanced interplay between these emotions and predictive behaviors. Despite encountering limitations related to data volume and resource availability, our study reports valuable discoveries concerning investor behavior and sentiment trends within the cryptocurrency market, informing strategic decision-making and future research endeavors.

**Keywords**
Cryptocurrency, Predictive statements,, Hope, Regret,

## 1. Introduction

A prediction is an assertion about future events based on current data, trends, or information. It involves leveraging historical data, identifying patterns, and using models to forecast future outcomes. Predictions are common in many fields, including weather forecasting, finance, sports, science, health, and beyond.

In 2008, an individual known by the pseudonym Satoshi Nakamoto published a seminal paper titled 'Bitcoin: a peer-to-peer electronic cash system,' marking a pivotal moment in financial history. This innovation succeeded where many others had failed, introducing Bitcoin as a decentralized electronic cash system, now universally known as cryptocurrency. As time progressed, numerous other cryptocurrencies emerged, capturing the interest of a diverse audience. Some individuals viewed these digital assets as investment opportunities, others used them for financial transactions, and many engaged in trading [1, 2]. Similar to our study, various research efforts have employed sentiment analysis to predict cryptocurrency prices, using data sources such as news [3, 4], tweets [5, 6], and other related datasets.

[*]Corresponding author.
✉ mshahikit2022@cic.ipn.mx (M. S. Tash)

The advancement of pre-training techniques has empowered large language models (LLMs) to master natural language processing (NLP) tasks with only a few examples provided "in context," eliminating the need to modify the model's parameters [7]. Few-shot learning, the capability to learn from a small number of examples, is a fundamental aspect of artificial intelligence [8].

In this study, we aim to identify predictive statements within cryptocurrency investment tweet data. We utilize a structured approach to classify these tweet segments into four categories: Predictive Incremental, Predictive Decremental, Predictive Neutral, or Non-Predictive, by leveraging the capabilities of GPT-4o. Additionally, we utilized two existing tasks, namely, hope speech detection and regret detection, to calculate the percentage of hope and regret expressed in these cryptocurrencies tweets.

Our comprehensive research explores the motivations and emotions of cryptocurrency investors. We analyze levels of hope, regret, and the prediction of cryptocurrency performance among users discussing digital currencies on X platform. For this purpose, we selected 1,000 tweets each for five cryptocurrencies—Cardano, Binance, Fantom, Matic, and Ripple—resulting in a dataset of 5,000 tweets. We employed few-shot learning with GPT-4o, the latest advanced model, to detect labels for these tweets. GPT-4o, with "o" standing for "omni," is a multimodal model capable of processing both text and image inputs and generating text outputs. It maintains the high intelligence of GPT-4 Turbo but operates more efficiently, producing text at twice the speed and half the cost.

Furthermore, we further explored the aspect of "reasons and significance." In simpler terms, our objective was to determine which cryptocurrency holds more promise, which carries more regret, which exhibits incremental predictability, and which demonstrates decremental predictability compared to other cryptocurrencies. To accomplish this, we examined the following contributions:

- A new model is introduced that provides predictive statements along with their categorization.
- Each comment is labeled according to its task: Predictive statement, Hope speech detection, or Regret detection.
- The percentage of hope, regret, and predictions for each coin is demonstrated.
- Various themes of hopes and regrets associated with each coin are identified.

## 2. Definitions

### 2.1. Predictive Statement

In this task, a prediction refers to a statement about the future performance or trend of an investment or market within tweets. We categorize these predictions as incremental (indicating expected improvement), decremental (indicating expected decline), or neutral (indicating no significant change). Our goal is to analyze investment-related tweets to determine which category—incremental, decremental, or neutral—has the highest percentage of predictions.

- **Incremental predictions** refer to forecasts that indicate a positive trend or enhancement in a future event or outcome. These predictions suggest that there will be growth, improvement, or an upward shift in the situation being analyzed. For instance, in business, an incremental prediction might forecast a rise in sales or an increase in market share.
- **Decremental predictions** pertain to forecasts that signal a negative trend or decline in a future event or outcome. These predictions indicate a reduction, deterioration, or downward movement in the scenario under consideration. For example, in economics, a decremental prediction could suggest a drop in GDP or a decline in employment rates.
- **Neutral predictions** describe forecasts that predict no significant change in a future event or outcome. These predictions suggest stability or stasis, where the current conditions are expected to remain relatively unchanged. In environmental studies, a neutral prediction might imply that current pollution levels will stay constant over a given period.
- **Non-predictive** text segments are those that do not contain any forecasts or projections about future events or outcomes. These segments may provide descriptions, explanations, or analyses that are focused on past or present conditions without making any statements about what might happen in the future. Examples of different categories of Predictive statements can be found in Table 1.

| Criteria | Examples |
|---|---|
| **Incremental** | There will be a steady increase in market share over the next quarter. |
| **Incremental** | Profits are expected to double by the end of the year. |
| **Incremental** | The company anticipates growth in revenue due to new product launches. |
| **Decremental** | Sales are projected to decline in the next fiscal quarter. |
| **Decremental** | There will be a 20% decrease in production efficiency. |
| **Decremental** | The market forecasts a decrease in consumer confidence. |
| **Neutral** | The company expects revenue to remain consistent in the upcoming quarter. |
| **Neutral** | There is uncertainty regarding future market conditions. |
| **Non-Predictive** | Blockchain technology is revolutionizing various industries worldwide. |
| **Non-Predictive** | The company reported record profits for the current fiscal year. |

**Table 1**
Examples of Different Categories of Predictive Statements

## 2.2. Hope Detection

Hope is a remarkable human ability that allows individuals to imagine future possibilities and their potential outcomes. These visions strongly influence emotions, actions, and mindset, despite the possibility of the desired outcome being improbable.[9, 10] In this research, we analyze various categories of hope within our data, including:

- **Generalized Hope** within the realm of cryptocurrency data embodies a pervasive sense of optimism and hopefulness that permeates the landscape. This sentiment isn't tied to any specific event or outcome but rather signifies a positive outlook on the overall dynamics, developments, and future possibilities in the cryptocurrency domain. This encompassing hope extends to anticipating price increases or decreases, aiming to leverage these fluctuations for more favorable investment opportunities, influencing one's optimistic stance within the cryptocurrency sphere [11, 12, 13, 14].

In the domain of cryptocurrency,

- **Not Hope** refers to the absence of any inclination toward hope, wishful thinking, desire for specific outcomes, or future-oriented expectations within the context of the discussed cryptocurrency-related content. Such tweet exhibits no indicators of positive anticipation, desired outcomes, or forward-looking expectations in this domain.

In the cryptocurrency realm,

- **Realistic Hope** involves envisioning outcomes that are reasonable, meaningful, and within the realms of possibility. This hopeful perspective encompasses expectations for sensible and likely results, often tethered to regular and expected events or developments within the cryptocurrency space. Notably, when Bitcoin, as a leader in cryptocurrencies, experiences an increase, it often sparks hope among traders for a positive transformation within the broader cryptomarket, influencing their realistic hopes for favorable changes [15, 16, 14].
- **Unrealistic Hope** within the cryptocurrency domain, represents an infrequent occurrence that occasionally emerges. It typically involves a desire for improbable events or outcomes to materialize, despite their extremely low or non-existent likelihood. This form of hope might arise from emotional states like anger, sadness, or depression, compelling individuals to anticipate unrealistic events or outcomes. For instance, an individual might fervently believe that purchasing a particular coin will instantly transform them into a wealthy individual, even when the likelihood of such an outcome is highly improbable. This hope lacks rational grounding and a meaningful basis within the context of cryptocurrencies [17, 18, 19].

## 2.3. Regret Detection

Regret [20] emerges as a negative feeling in reaction to situations or events that individual desires had unfolded in an alternative manner [21]. Frequently, it's linked with sentiments of remorse, disillusionment, and self-criticism [22, 23]. This particular task aims to categorize texts into three distinct groups: 'Regret due to taking action' (Action), 'Regret due to not taking action' (Inaction), and 'No expressed regret'.

- **Regret by Action**
  In the cryptocurrency domain, "Regret by Action" encompasses the feeling of remorse triggered by a past action, such as buying a particular coin, that leads to unfavorable outcomes, notably, a decrease in its value. This form of regret surfaces when a decision or action results in undesired consequences, specifically within the context of cryptocurrency investments, impacting an individual's financial holdings or expectations.
- **Regret by Inaction**
  In the realm of cryptocurrency, "Regret by Inaction" materializes when an individual refrains from taking a specific action, such as buying a coin, and later experiences remorse due to the subsequent increase in its value. This type of regret stems from the decision to not act, leading to a sense of regret or missed opportunity caused by the favorable outcomes that follow the action not taken within the cryptocurrency domain.
- **No Regret**
The text does not express any form of regret.

## 3. Literature Review

### 3.1. Cryptocurrency

Huang et al. [24] focus on forecasting cryptocurrency price volatility by assessing social media sentiment and establishing correlations. Utilizing the prominent Chinese platform Sina-Weibo, a comprehensive corpus was gathered, comprising 24,000 tweets and 70,000 associated comments related to Bitcoin, ETH, or XPR over an eight-day span. A specialized sentiment dictionary was formulated, and a long short-term memory (LSTM) based recurrent neural network, combined with historical cryptocurrency price data, was proposed to predict future price trends. Comparative experiments showcased the proposed method's superiority, the precision and recall of the Auto Regression model were 73.4% and 80.2%, respectively, while the LSTM Sentiment Analyzer achieved 87.0% precision and 92.5% recall.

Nasekin and Chen [25] examine the market sentiment among cryptocurrency investors and traders using the StockTwits platform. Employing machine learning techniques, specifically, recurrent neural networks (RNNs), sentiment indices are constructed to capture the evolving opinions of the cryptocurrency community regarding the market over time. These newly created sentiment indices are then integrated into predictive models for the autoregressive mean and variance of returns in the CRIX cryptocurrency index. Analysis reveals that the RNN LSTM configuration, coupled with a pre-trained embedding layer, demonstrates superior predictive performance compared to other RNN setups using GRU or LSTM units, either with pre-trained or randomly initiated embeddings. The study utilizes a unique dataset comprising 1.22 million messages associated with 425 cryptocurrencies, sourced from the microblogging platform StockTwits[1], covering the period from March 2013 to May 2018.

Shahiki Tash et al. [26] contribute a comprehensive analysis of English tweets associated with cryptocurrencies, focusing on a curated dataset encompassing nine digital coins. Leveraging tools like LIWC (Linguistic Inquiry and Word Count) [27], Emotion [2], Sentiment [28], and Readability [29, 30, 31, 32] analysis, the study scrutinizes linguistic attributes and sentiment trends within these tweets. Notably, the examination reveals distinct linguistic traits across various cryptocurrencies, with a prevailing tendency towards logical and formal thought in tweets. Fantom coin emerges as prominently characterized by these features, showcasing discussions leaning towards logical and formal thinking in comparison to other coins. Moreover, the study highlights differential sentiment scores among cryptocurrencies, notably observing Fantom coin generating higher positive sentiment while Ripple coin exhibits elevated negative sentiment. Emotionally, most cryptocurrencies display a balanced profile, emphasizing "Anticipation," with Dogecoin prominently associated with this emotional category. Readability metrics imply varying levels of comprehension challenges, with Dogecoin tweets posing higher readability complexities and Ethereum demanding a higher reading proficiency, indicative of differing communication styles among these cryptocurrencies.

---

[1]https://stocktwits.com/
[2]https://textblob.readthedocs.io/en/dev/

The study conducted by Ibrahim [33] focus on predicting initial movements in the Bitcoin market by utilizing sentiment analysis of Twitter data Pano and Kashef [34]. Their primary aim was to introduce a Composite Ensemble Prediction Model (CEPM) constructed using sentiment analysis. They applied a blend of data mining techniques, machine learning algorithms, and natural language processing to interpret public sentiment and mood related to cryptocurrencies. The study assessed various models, including Logistic Regression, Binary Classified Vector Prediction, Support Vector Mechanism, Naïve Bayes, and a standalone XGBoost Wang et al.,[35] for sentiment analysis. Notably, the CEPM outperformed other methods, underscoring its efficacy in predicting early movements in the Bitcoin market by leveraging sentiment analysis of Twitter data.

Dag et al. [36] presents a semi-explanatory predictive framework tailored for BTC trading choices, leveraging an extensive feature selection process. Utilizing a dataset centered on factors pertinent to BTC price changes and drawing from prior studies [37, 38, 39, 40], the proposed Tree Augmented Naïve (TAN) model undergoes validation via 10-fold cross-validation. The outcomes reveal an average AUC of 0.652, sensitivity of 0.697, specificity of 0.641, and accuracy of 0.667, leveraging just 6 variables.

## 3.2. Hope and Regret Detection

Prior research efforts aimed at tackling hate Shahiki-Tash et al. [41] in English, disregarding the wider scope of harmful content across languages. To bridge this gap, a fresh Hope Speech dataset for Equality, Diversity, and Inclusion (HopeEDI) Chakravarthi [42] was compiled, featuring user comments from YouTube in English, Tamil, and Malayalam. This dataset, totaling 28,451, 20,198, and 10,705 comments respectively, underwent meticulous annotation for hope speech detection. In evaluating dataset reliability, Krippendorff's alpha Krippendorff [43] was used for inter-annotator agreement. Notably, the SVM classifier demonstrated the lowest macro-average F1-scores within the HopeEDI dataset, while differing model performances were observed. Decision trees performed better in English and Malayalam, while logistic regression excelled in the Tamil subset.

Balouchzahi et al. [44], present a fresh dataset for categorizing tweets as "Hope" or "Not Hope," further classifying them into "Generalized Hope," "Realistic Hope," and "Unrealistic Hope." Derived from English tweets from the first half of 2022, the dataset was collected using Tweepy's API, filtering tweets on various parameters, focusing on domains such as women's child abortion rights, black people's rights, religion, and politics. The collection involved approximately 100,000 tweets, from which a randomly chosen subset of 10,000 tweets was used for further analysis. Notable F1-scores were observed in model performance: LR achieved 0.80 in binary hope speech detection, CatBoost scored 0.54 in multiclass hope speech detection, BiLSTM attained 0.79 for binary hope speech detection and 0.57 for multiclass detection, and bert-base-uncased showed promising F1-scores of 0.85 for binary hope speech detection and 0.72 for multiclass hope speech detection, marking the best performance outcomes.

The study Balouchzahi et al. [20] works on regret expression on social media, presenting a new Reddit text dataset categorized into Regret by Action, Regret by Inaction, and No Regret. Focused on regret detection and domain identification, the dataset compiled posts from "regret," "regretful parents," and "confession" subreddits via the Pushshift API and PMAW framework. The obtained posts were filtered and duplicates were removed resulting in 3440 posts. The study assessed dataset reliability using baseline models and state-of-the-art CNN and BiLSTM models. Moderate F1-scores were achieved, with CNN scoring 0.612 for Regret Detection (text), 0.715 for Regret Detection (text + title), and 0.58 for Domain Identification (text), while BiLSTM achieved 0.629 for Domain Identification (text + title), highlighting their performance in regret detection and domain classification.

Balouchzahi et al. [45] explore a deep-learning method that incorporates linguistic and psycholinguistic traits for hope speech identification. Their dataset [46] contains English YouTube comments and Spanish Tweets. They evaluate their proposed model using macro F1-scores, treating all classes equally, and weighted F1-scores, accounting for class distribution. Compared to other models, their proposal excelled, achieving a weighted F1-score of 0.870 in English, emphasizing LIWC features. For Spanish, the amalgamation of LIWC, word, and character n-grams attained a weighted F1-score of 0.790. These results indicate that uncomplicated deep learning models, integrating linguistic and psycholinguistic components, can outperform intricate language models.

## 4. Methodologhy

The proposed methodology introduces a fresh approach termed Predictive statement task, that categorizes predictive statements into four distinct labels: Incremental, Decremental, Neutral, and Non-Predictive. This method is specifically tailored for analyzing comments pertaining to cryptocurrencies, with a primary focus on identifying predictive statements. Additionally, we incorporated two pre-existing tasks: expressions of regret and hope speech. Following this, we embarked on classifying data associated with these tasks. Leveraging few-shot learning, the GPT-4o model underwent training using labeled examples to achieve precise comment classification. Regret detection entailed labeling comments with RegretType, making distinctions between Regret by Action, Regret by Inaction, and No Regret, also employing few-shot learning. Likewise, hope speech detection involved categorizing comments into Generalized Hope, Realistic Hope, Unrealistic Hope, or Not Hope, utilizing the same few-shot learning method. The workflow encompassed stages such as data collection, preprocessing, model training with few-shot learning, and classification into specific labels.

The following is a detailed description of each stage.

### 4.1. Data Collection

The information used for our analysis was sourced from two research papers by [26, 47], which collected data from the X platform spanning September 2021 to March 2023. Initially, the dataset included 115,899 tweets, from which a subset of 5,000 comments was chosen for examination. From these papers, a random selection of 1,000 comments for each of the cryptocurrencies (Cardano, Binance, Fantom, Matic, Ripple) was made. Subsequently, 1,000 tweets were selected for each cryptocurrency, and using a few-shot learning-based GPT-4o model, labels were assigned.

## 4.2. Data Evaluation

Inter-annotator agreement (IAA) is used to evaluate the extent of consensus among annotators. For our task evaluation, the GPT-4o model was assessed by comparing it to manual annotations using Cohen's Kappa Coefficient scores. A random sample of 1,000 comments was selected, and the following results were obtained: 0.4393% for Hope detection, 0.5796% for Regret detection, and 0.7173% for Predictive statement detection. The robustness of the datasets and the thoroughness of the annotation process are highlighted by these scores.

## 4.3. Data Preprocessing

After acquiring the dataset, we initiated a multi-step data preprocessing protocol aimed at refining and optimizing the data. This process comprised the following primary stages:

**URL Removal**: Employing a regular expression pattern, we systematically identified and eliminated any URLs present within the dataset.

**Text Cleaning**: This phase involved the systematic removal of special characters, such as punctuation marks, leveraging a dedicated dictionary of such characters. Additionally, words with a length equal to or less than two characters were excluded. The outcome was a refined version of the textual data devoid of unnecessary elements.

For each comment, we implemented preprocessing procedures to ensure its compatibility for subsequent analysis, encompassing:

- Elimination of superfluous characters or noise.
- Tokenization of comments to facilitate improved handling by the model.
- Normalization of text to ensure uniformity and consistency throughout the dataset.

## 4.4. Model Training and Execution

In the model training and execution phase, the GPT-4o model is fine-tuned with the provided examples using few-shot learning. For each comment, the structured prompt containing the comment and example labels is fed into the model. The input is processed by the model, and a classification label is generated based on its understanding of the comment's sentiment and predictive nature. This classification label is then stored in the respective columns (PredictionType, RegretType, HopeType) of the dataset, facilitating easy retrieval and analysis of the model's classifications. This systematic approach ensures the effective training and utilization of the GPT-4o model for accurate sentiment analysis and prediction categorization within cryptocurrency-related discussions. More details can be found in Table 2.

| Category | Comment | Label | Prompt Description |
|---|---|---|---|
| Predictive | Profits are expected to double by the end of the year. | Predictive Incremental | You are an assistant trained to categorize comments into Predictive Incremental, Predictive Decremental, Predictive Neutral, or Non-Predictive. Just give us one label per row. |
| | Due to the recent economic downturn, we anticipate a decrease in consumer spending next quarter. | Predictive Decremental | |
| | The company expects revenue to remain consistent in the upcoming quarter or There is uncertainty regarding future market conditions. | Predictive Neutral | |
| | Blockchain technology is revolutionizing various industries worldwide. | Non-Predictive | |
| Regret | I regret buying that coin, it's lost so much value. | Regret by Action | You are an assistant trained to categorize comments about cryptocurrencies into three types of regret. Just give us one label per row: Regret by Action, Regret by Inaction, and No Regret. |
| | I should have bought that coin when it was cheaper, now it's too late. | Regret by Inaction | |
| | I'm glad I didn't invest in that coin, it's crashing. | No Regret | |
| Hope | Excited about the future of cryptocurrencies! The innovation and potential in this space are truly remarkable. | Generalized Hope | You are an assistant trained to categorize comments about cryptocurrencies into four types of hope. Just give us one label per row: Generalized Hope, Not Hope, Realistic Hope, and Unrealistic Hope. |
| | I doubt this coin will ever increase in value. | Not Hope | |
| | With the recent trends, it's likely that Bitcoin will hit a new high. | Realistic Hope | |
| | I'm sure this tiny investment will make me a millionaire overnight. | Unrealistic Hope | |

**Table 2**
Examples and Prompts for Categorizing Cryptocurrency Comments

## 5. Results and Analysis

### 5.1. Prediction

Examining the data, it's notable that each cryptocurrency has a significant portion of predictive tweets, with varying degrees of incremental and decremental predictions.

Among the predictive categories, Cardano's tweets are fairly evenly distributed between incremental (2.2%) and decremental (2.3%) predictions, suggesting a balanced sentiment regarding its future performance. However, the proportion of neutral predictions is minimal (0.1%), indicating a lack of consensus on its future trajectory.

Similar to Cardano, Binance also has a relatively balanced distribution of predictive tweets, with a slightly higher percentage of decremental predictions (4.5%) compared to incremental predictions (4.4%). Again, neutral predictions are minimal (0.2%), indicating a clear directional sentiment among users discussing Binance.

Matic stands out with a notably higher percentage of incremental predictions (8.1%) compared to decremental predictions (2.4%). This suggests a more optimistic sentiment among users discussing Matic, anticipating growth or improvement in its future performance. However, similar to other coins, neutral predictions remain minimal (0.3%).

Fantom's predictive tweets also show a balanced distribution between incremental (3.6%) and decremental (1.9%) predictions, albeit with a slightly higher emphasis on incremental predictions. As with other coins, neutral predictions are very rare (0.1%).

Ripple's predictive tweets are somewhat evenly distributed between incremental (3.7%) and decremental (4.3%) predictions, indicating a mixed sentiment regarding its future performance. Again, neutral predictions are minimal (0.4%), reflecting a clear directional sentiment among users discussing Ripple. Further details can be found in Table 3.

In summary, while there's a mix of sentiment expressed in predictive tweets across all cryptocurrencies, Matic stands out with a notably higher proportion of incremental predictions, suggesting a more positive outlook compared to other coins. The summary of the distribution in Figure 1 is as follows:

| Coins | Non-Predictive | Predictive Decremental | Predictive Incremental | Predictive Neutral |
|---|---|---|---|---|
| **Cardano** | 95.4% | 2.3% | 2.2% | 0.1% |
| **Binance** | 90.9% | 4.5% | 4.4% | 0.2% |
| **Matic** | 89.2% | 2.4% | 8.1% | 0.3% |
| **Fantom** | 94.4% | 1.9% | 3.6% | 0.1% |
| **Ripple** | 91.6% | 4.3% | 3.7% | 0.4% |

**Table 3**
Percentage distribution of predictive statements across various cryptocurrencies

### 5.2. Hope Detection

The analysis of sentiment distribution across various cryptocurrencies reveals intriguing patterns in how users perceive and express sentiments about each digital asset. Fantom emerges as the cryptocurrency garnering the highest percentage of tweets categorized as "Not-Hope," indicative of a prevalent sentiment of skepticism or negativity surrounding this particular asset. Conversely, Matic exhibits a significant surge in tweets categorized as "Unrealistic Hope," suggesting an abundance of optimism or unrealistic expectations associated with this cryptocurrency. These divergent sentiment distributions underscore the nuanced nature of sentiment dynamics within the cryptocurrency space, where each asset engenders a unique sentiment landscape shaped by factors such as market performance, technological advancements, and community sentiment. One can find more details in Table 4.

Furthermore, the varying degrees of "Generalized Hope" and "Realistic Hope" across different cryptocurrencies highlight the multifaceted nature of sentiment expression within the cryptocommunity. While some assets like Binance may evoke more realistic expectations or generalized optimism, others like Ripple may witness a higher prevalence of unrealistic expectations or speculative fervor. Understanding these nuanced sentiment dynamics is crucial for investors, analysts, and stakeholders in navigating the complex landscape of cryptocurrency markets, where sentiment plays a pivotal role in shaping market trends and investor behavior. By looking deeper into these sentiment distributions, stakeholders can glean valuable insights into market sentiment trends and make informed decisions regarding investment strategies and market positioning. The distribution of different types of hope can be found in Figure 2.

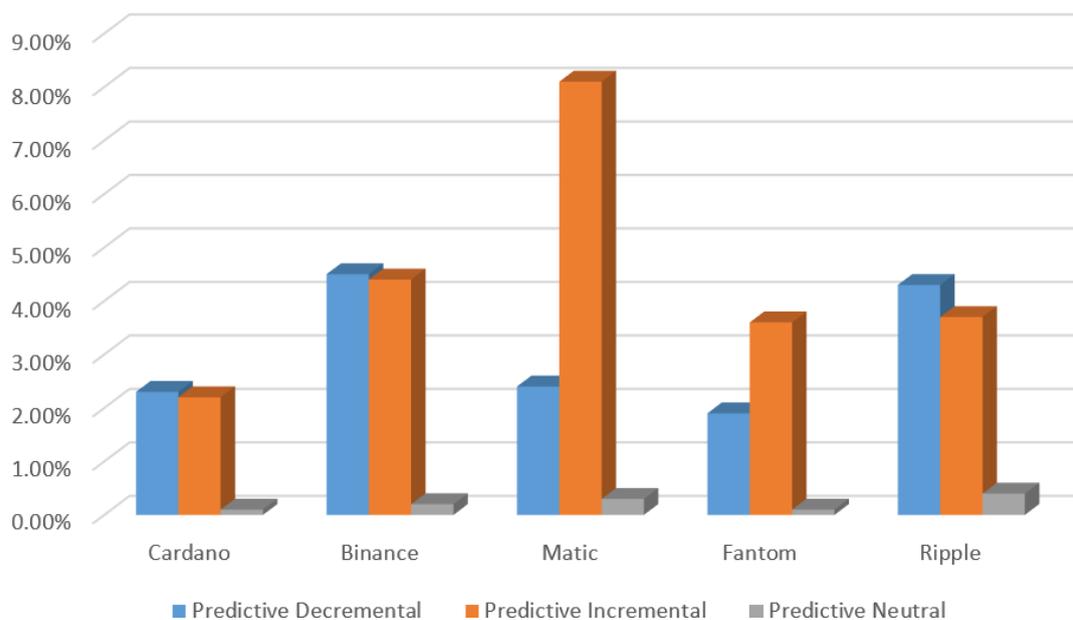

**Figure 1:** The percentage of predictive statements for cryptocurrency

|  | Not-Hope | Unrealistic Hope | Generalized Hope | Realistic Hope |
|---|---|---|---|---|
| **Cardano** | 76% | 9.8% | 7.6% | 6.6% |
| **Binance** | 75.1% | 18.4% | 2.7% | 3.8% |
| **Matic** | 63.2% | 23.4% | 2.9% | 10.5% |
| **Fantom** | 80.5% | 15.8% | 0.7% | 3% |
| **Ripple** | 72.6% | 21.3% | 4.8% | 1.3% |

**Table 4**
Distribution of Hope Categories Across Different Cryptocurrencies

## 5.3. Regret Detection

The analysis of regretful sentiment distribution across various cryptocurrencies unveils distinct patterns in user perceptions and experiences within the cryptosphere. Fantom and Binance stand out with the lowest percentages of tweets categorized as "Regret by Action" and "Regret by Inaction." This suggests a prevailing sentiment of contentment or confidence among users discussing these assets, possibly due to positive market performance or successful project developments. Conversely, Ripple exhibits the highest percentage of tweets categorized as "Regret by Action," indicating a notable prevalence of regrets stemming from actions taken in relation to Ripple, such as buying or selling decisions. This could reflect challenges or uncertainties faced by Ripple during the analyzed period, leading to a higher proportion of users expressing regret or dissatisfaction with their interactions or investments in the cryptocurrency. Matic demonstrates a relatively higher percentage of tweets categorized as "Regret by Inaction," hinting at potential missed opportunities or decisions not taken by users in relation to Matic. Additional details can be found in Table 5. These nuanced variations underscore the diverse nature of regretful sentiments within the cryptocurrency space, shaped by factors such as market dynamics, project developments, and individual investment strategies. One possible reason for the lowest percentage of different types of regret comments across all cryptocurrencies could be attributed to the overall positive sentiment prevailing in the cryptocurrency space during the time period analyzed. Additionally, users may be more inclined to share positive experiences or optimistic outlooks rather than dwell on past regrets, leading to a lower proportion of regretful sentiments in the dataset. Moreover, the lower percentages of regretful sentiments could also reflect the relative stability or positive performance of these cryptocurrencies during the analyzed period, mitigating instances of regret among users. A comparison between different types of regret can be found in Figure 3.

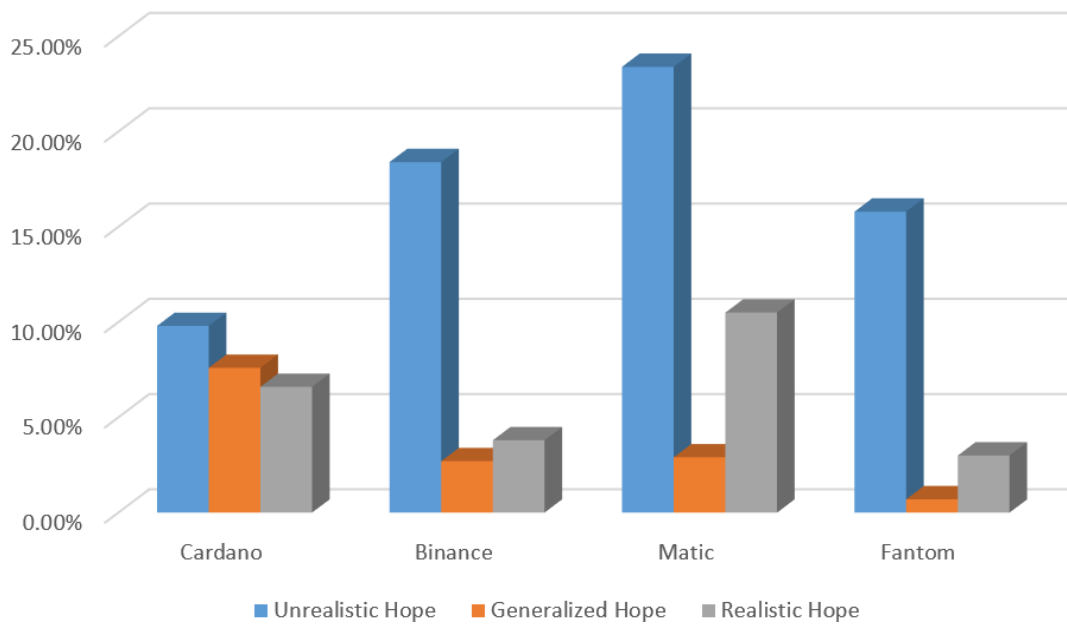

**Figure 2:** Compare the distribution of different types of hope

## 6. Discussion

Analyzing both hope, regret, and predictive sentiments in cryptocurrency-related discussions can yield profound insights into investor sentiment and market dynamics. The data suggests that predictive sentiments, such as "Predictive Incremental" or "Predictive Decremental," significantly influence the overall emotional landscape. For instance, a higher prevalence of "Predictive

|  | No Regret | Regret by Action | Regret by Inaction |
|---|---|---|---|
| **Cardano** | 97.8% | 1.9% | 0.3% |
| **Binance** | 98.1% | 1.4% | 0.5% |
| **Matic** | 97.6% | 1.4% | 1.0% |
| **Fantom** | 98.2% | 1.3% | 0.5% |
| **Ripple** | 96.2% | 2.8% | 1.0% |

**Table 5**
Distribution of Regret Categories Across Different Cryptocurrencies

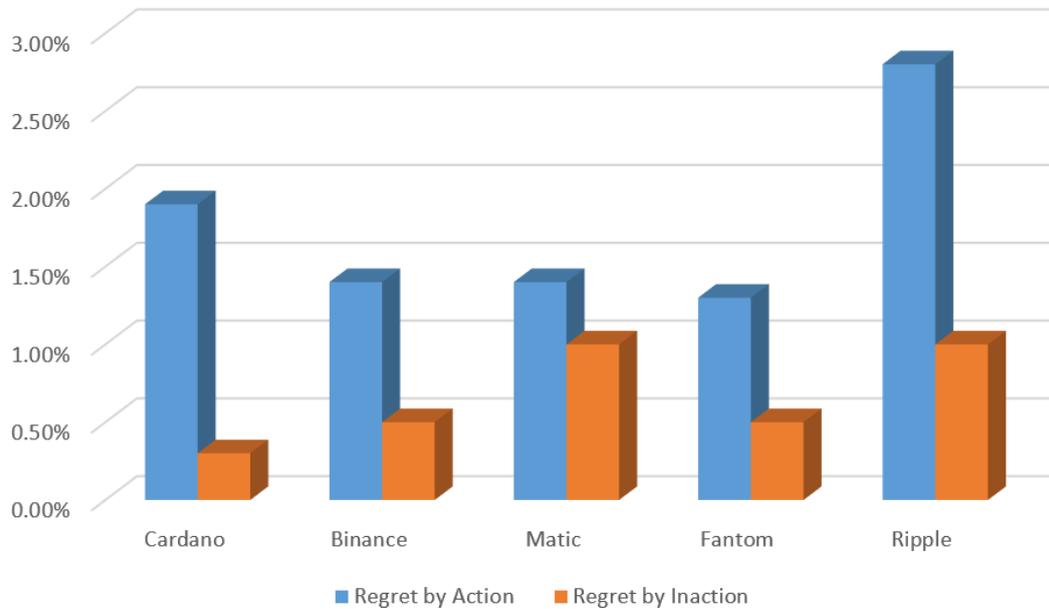

**Figure 3:** To compare the percentages of regret by action and regret by inaction:

Incremental" tweets may reflect investor optimism about future price movements, potentially correlating with elevated levels of hope and diminished regret. Conversely, an abundance of "Predictive Decremental" tweets may indicate a more cautious outlook, potentially aligning with heightened regret and reduced hope. Understanding these predictive sentiments alongside hope and regret provides a holistic view of investor behavior and sentiment trends in the cryptocurrency market, empowering stakeholders to make more informed decisions regarding investment strategies and risk management.

Furthermore, exploring the relationship between hope and regret uncovers an intriguing inverse correlation. Higher instances of "Unrealistic Hope" tweets may coincide with lower occurrences of "Regret by Action" or "Regret by Inaction," suggesting that exaggerated optimism could mitigate feelings of regret among investors. Conversely, a preponderance of "Regret by Action" or "Regret by Inaction" tweets may dampen sentiments of hope, as regretful reflections on past decisions overshadow optimistic outlooks for the future. This dynamic underscores the nuanced interplay between emotions in shaping investor perceptions and decision-making processes within the cryptocurrency market.

In conclusion, a comprehensive analysis of hope, regret, and predictive sentiments provides invaluable insights into investor behavior and sentiment trends in the cryptocurrency domain. By understanding the intricate dynamics between these emotions, stakeholders can navigate market fluctuations more effectively and strategically position themselves for success. This nuanced understanding enhances the ability to interpret sentiment trends accurately and make data-driven decisions, ultimately maximizing opportunities and mitigating risks in the cryptocurrency market.

## 7. Conclusion and Future Work

In this study, we examined five prominent cryptocurrencies, namely Cardano, Binance, Matic, Fantom, and Ripple, comprising a total dataset of 5,000 comments, with each cryptocurrency represented by 1,000 comments. Introducing a novel classification scheme named "Prediction," we categorized comments into four distinct categories: Predictive Incremental, Predictive Decremental, Predictive Neutral, or Non-Predictive. Leveraging GPT-4o, we performed classification tasks, observing a notable trend in the Matic coin, where users exhibited a higher propensity for prediction, particularly in the Predictive Incremental category, indicating a more favorable sentiment towards investment in this coin. Conversely, Cardano demonstrated the lowest percentage of prediction, suggesting comparatively lower enthusiasm for investment. Notably, Binance displayed a higher percentage of Predictive Decremental sentiment, suggesting a prevailing negative outlook toward investment in this particular cryptocurrency. Subsequent to prediction analysis, we pursued two ancillary tasks: hope speech detection and regret detection. Within the hope category, Matic exhibited the highest proportion of hope sentiments, predominantly characterized by Unrealistic Hope, while also demonstrating the highest percentage of Realistic Hope. Conversely, Cardano displayed the highest proportion of General Hope sentiments, implying a more optimistic outlook regarding future investment prospects in Matic. As a matter of fact, we encountered challenges in conducting a detailed regret analysis due to the notably low occurrence of regret-labeled comments. For future research, we intend to expand the dataset size and assess the correlation between cryptocurrency price fluctuations and sentiment analysis outcomes at the data collection juncture. Additionally, we aim to ascertain the accuracy percentage of GPT-4o predictions, thereby enhancing the robustness and reliability of our analysis.

## 8. Limitation

A primary limitation of this study pertained to the insufficient volume of data available, which constrained our ability to conduct a more comprehensive analysis, particularly in the prediction aspect. The scarcity of resources hindered our capacity to augment the dataset adequately, thereby limiting the scope and depth of our predictive analysis.

## Acknowledgements

The work was done with partial support from the Mexican Government through the grant A1-S-47854 of CONACYT, Mexico, grants 20241816, 20241819, and 20240951 of the Secretaría de Investigación y Posgrado of the Instituto Politécnico Nacional, Mexico. The authors thank the CONACYT for the computing resources brought to them through the Plataforma de Aprendizaje Profundo para Tecnologías del Lenguaje of the Laboratorio de Supercómputo of the INAOE, Mexico and acknowledge the support of Microsoft through the Microsoft Latin America PhD Award.